\documentclass{article} % For LaTeX2e
\usepackage{iclr2021_conference,times}

\usepackage{amsmath,amsfonts,bm}

\def\eqref#1{equation~\ref{#1}}
\def\1{\bm{1}}

\DeclareMathAlphabet{\mathsfit}{\encodingdefault}{\sfdefault}{m}{sl}
\SetMathAlphabet{\mathsfit}{bold}{\encodingdefault}{\sfdefault}{bx}{n}

\usepackage{bbm}
\usepackage{hyperref}
\usepackage{url}
\usepackage{booktabs}
\usepackage{tabularx}
\usepackage{multirow}
\usepackage{graphicx}
\usepackage{tikz}
\usepackage{amssymb}
\usepackage{pifont}
\usepackage{xcolor}
\usepackage{soul}
\usepackage{wrapfig}
\sethlcolor{yellow}

\usepackage{subcaption}
\usepackage{graphicx}

\usepackage{lipsum}

\newcommand{\ie}[0]{\emph{i.e., }}

\newcommand{\eg}[0]{\emph{e.g., }}

\newcommand{\RN}[1]{%
	\textup{\lowercase\expandafter{\it \romannumeral#1}}%
}

\title{Stabilizing RLHF with Advantage Model and Selective Rehearsal}
\title{Stabilizing RLHF through Advantage Model and Selective Rehearsal}

\author{Baolin Peng\thanks{Equal Contribution}, Linfeng Song\textsuperscript{$*$}, Ye Tian, Lifeng Jin, Haitao Mi, Dong Yu\\
Tencent AI Lab\\
\texttt{\{baolinpeng,lfsong,yaptian,lifengjin,haitaomi\}@global.tencent.com}  \\
\\
}

\iclrfinalcopy % Uncomment for camera-ready version, but NOT for submission.
\begin{document}

\maketitle

\begin{abstract}

Large Language Models (LLMs) have revolutionized natural language processing, yet aligning these models with human values and preferences using RLHF remains a significant challenge. This challenge is characterized by various instabilities, such as reward hacking and catastrophic forgetting. In this technical report, we propose two innovations to stabilize RLHF training: (\RN{1}) \textit{Advantage Model}, which directly models advantage score \ie extra reward compared to the expected rewards and regulates score distributions across tasks to prevent reward hacking. (\RN{2}) \textit{Selective Rehearsal}, which mitigates catastrophic forgetting by strategically selecting data for PPO training and knowledge rehearsing. Our experimental analysis on public and proprietary datasets reveals that the proposed methods not only increase stability in RLHF training but also achieve higher reward scores and win rates\footnote{Work in progress}.

\end{abstract}

\section{Introduction}
\label{sec:introduction}

Large language models (LLMs) have become a fundamental element in advancing natural language processing (NLP) and artificial intelligence (AI), showcasing an impressive ability to generate text that is both semantically and contextually relevant~\citep{openai2023gpt,kopf2023openassistant,llama2}. 
Despite these advancements, LLMs have the risk of engaging in undesirable behaviors, such as fabricating information or producing biased, toxic, or even dangerous content, since LLMs are trained on a wide array of data, which can include low-quality sources. 
This has highlighted the necessities of LLM Alignments with human values, intentions, and preferences~\citep{brown2020language,ouyang2022training,anthropic,glaese2022improving}. 

% LLM Alignments pertains to the procedure of ensuring that LLMs not only effectively enhance surrogate training objectives but also ensure that their predictions align with their intended use and comply with ethical and safety guidelines established by humans.

Many approaches have been put forth to address the challenge LLM Alignments~\citep{anthropic,openai2023gpt,askell2021general}. Among these approaches, Reinforcement Learning from Human Feedback (RLHF) has demonstrated its efficacy in aligning language models with human preferences.
RLHF serves as a key component of training SoTA LLMs including exemplars such as OpenAI's GPT-4~\citep{openai2023gpt}, Anthropic's Claude~\citep{anthropic}, Google's Sparrow~\citep{glaese2022improving}, Bard, and Meta's Llama 2-Chat~\citep{llama2}.
RLHF elevates the capabilities of LLMs beyond the mere modeling of the distribution of their training data. It endows LLMs with the capacity to adapt their text generation distribution in a manner that are preferred by humans.

\begin{figure}[htbp]
    \centering
    \begin{subfigure}[b]{0.43\textwidth}
        \centering
        \includegraphics[width=\textwidth]{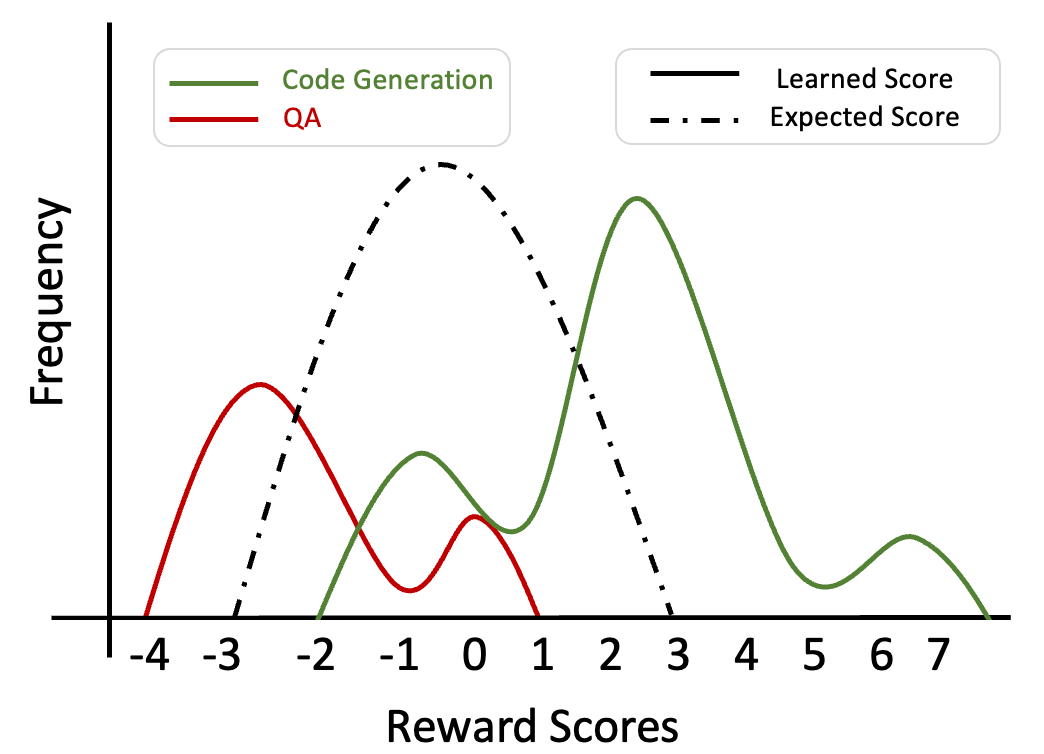}
        \caption{Reward score distributions.}
        \label{fig:running_examples_a}
    \end{subfigure}
    \quad
    \begin{subfigure}[b]{0.49\textwidth}
        \centering
        \includegraphics[width=\textwidth]{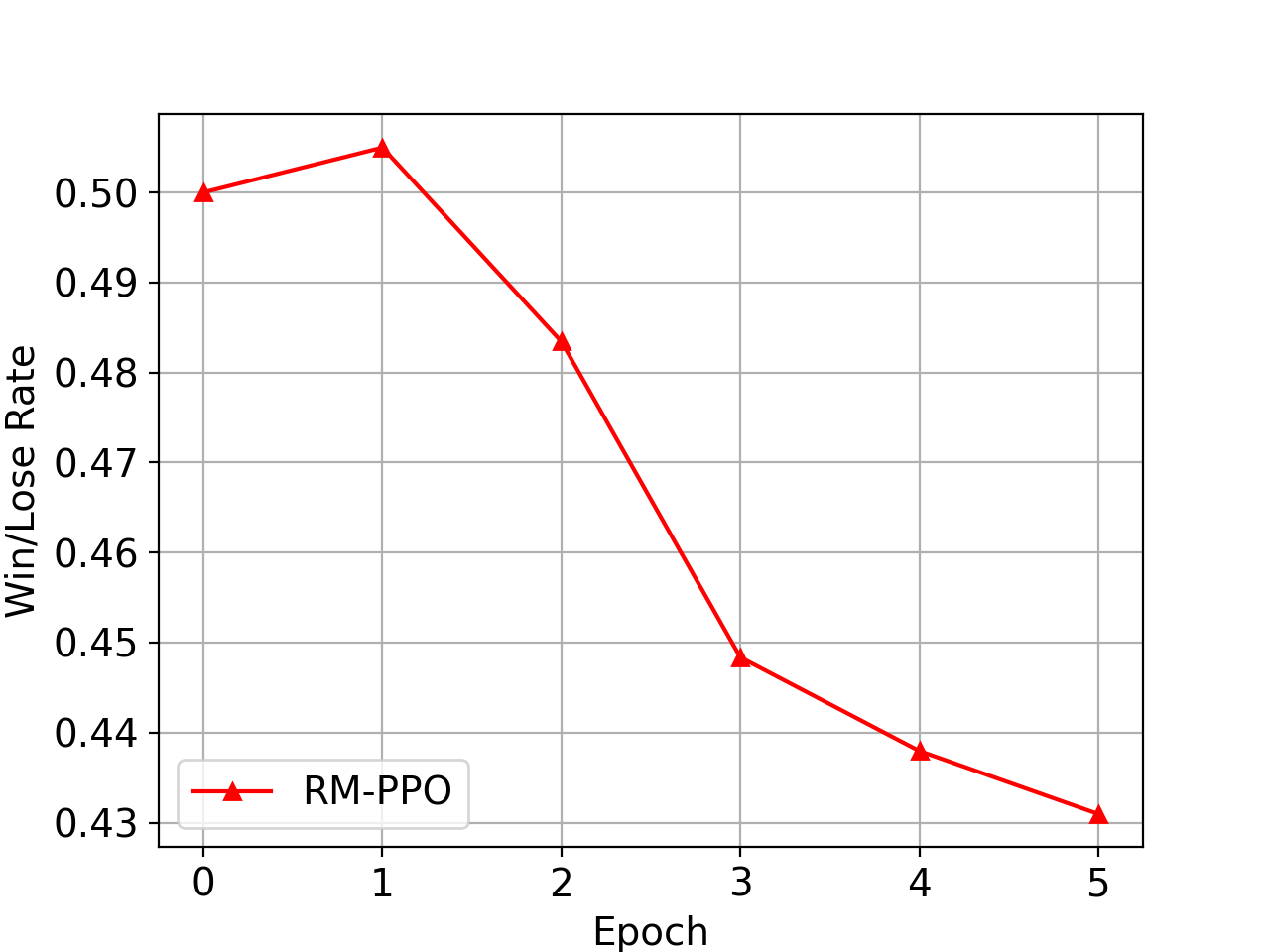}
        \caption{Win rate over the SFT model on the \textit{forget set} evaluated by GPT-4.}
        \label{fig:running_examples_b}
    \end{subfigure}
    \caption{\textit{Left}: The distribution of reward scores for both the QA and Code Generation tasks. There is a noticeable disparity in the learned reward score distributions between the two tasks, despite the expectation that the distributions should be similar. \textit{Right}: The win/loss rate over the SFT model on the forget set exhibits a significant decline. This drop in the win rate can be attributed to reward hacking and the phenomenon of catastrophic forgetting.}
    \label{fig:running_examples}
\end{figure}

However, training LLMs using RLHF is undoubtedly challenging, which demands an accurate and reliable reward model that approximates human judges, and a robust PPO algorithm for sustained policy improvements. Even with meticulous configurations, \textit{instabilities}, \eg gibberish responses (but high-reward)~\citep{stiennon2020learning,skalse2022defining}, 
forgetting learned knowledge, are usually observed during training, which leads to recurring failures. These instabilities have several causes:  (\RN{1}) different reward score distributions are learned for various categories by the reward model, potentially leading to \textit{reward hacking} issues~\citep{skalse2022defining}, a phenomenon where the model finds unintended ways to maximize the reward.  As depicted in Figure~\ref{fig:running_examples_a}, the reward model learns noticeable disparity in reward score distributions for Code Generation and QA tasks, 2 out of 61 tasks present in the preference data. 
% Q: What's the distributions of all examples in our preference data? 
Even with reward score normalizations, the fluctuating means and variances can induce unexpected model behaviors, such as transferring the response patterns of Code Generations to QA examples due to the higher reward scores. (\RN{2}) over-optimizing with PPO on examples that were well-aligned with humans in the Supervised Fine-Tuning (SFT) stage triggers \textit{catastrophic forgetting} issues~\citep{mccloskey1989catastrophic, gupta2023continual,khetarpal2022towards}. Models tend to overlook what was learned during the SFT stage, \ie PPO model underperforms the SFT model on expert-aligned examples~\footnote{Expert-aligned Examples are data samples that meet the standards and criteria delineated by experts and closely align with human preferences. These examples are used for SFT model training and evaluation.}, as shown in Figure~\ref{fig:running_examples_b}.
% expert-aligned examples.
% Q: missing another reference, what's the defination of `expert-aligned examples`?

Accordingly, in this technical report, we introduce two techniques to enhance the stability and effectiveness of the training of RLHF. Firstly, we propose \textit{Advantage Model} to balance the reward score distributions across various categories, thus averting the reward hacking dilemma that is often induced by noticeable differences score distributions. This is achieved by directly modeling the advantage score, \ie the extra reward one response can obtain compared with the expected reward, and regulating the advantage score distribution dynamically during training, ensuring that the variances and means are maintained within a reasonable range. Secondly, we introduce the \textit{Selective Rehearsal} to alleviate the catastrophic forgetting issue. We posit that not all data should be optimized equally in PPO training. As such, we propose a robust and effective data selector that automatically identifies what examples could be utilized for PPO training and should be used to rehearsal knowledge accumulated in the SFT stage, preventing the depreciation of the model’s performance on expert-aligned examples over time. Experiments on both public and proprietary data have demonstrated that our Advantage Model successfully balances reward score distributions across various examples while preserves ranking precision, and guide PPO training to achieve a higher reward score and win rate compared to the SFT model. Furthermore, Selective Rehearsal is able to avoid over-optimizing by selecting the most suitable examples for PPO training, thereby sustaining the performance on expert-aligned examples.

Our contributions are summarized as follows:
\begin{itemize}
\item We analyze and identify several causes of instability in RLHF training, namely, imbalanced learned reward score distributions and over-optimization of certain PPO training data, which lead to reward hacking and catastrophic forgetting issues.

\item We introduce the \textit{Advantage Model} to balance reward score distributions across various categories, and the \textit{Selective Rehearsal} strategy to discern which examples should be used for PPO training and which should be reserved for rehearsing knowledge accrued in the SFT stage. 
% These approaches are designed to alleviate reward hacking and catastrophic forgetting phenomena, respectively.

\item Through extensive experiments on both public and proprietary datasets, we demonstrate that the \textit{Advantage Model} and \textit{Selective Rehearsal} are able to stabilize RLHF training, achieving higher reward scores and win rates. %compared to the SFT models.
\end{itemize}

% - LLMs are powerful

% - RLFH is effective way for scalble Alignment

% - Training is instable (cite fudan).

% - There are several attempts to improve training stability, eg, even though effective, these methods did not solve the problem from root

% - we obsever two issues, 1. rm scores are sharply different, 2. over optimizing tasks that are already aligned during SFT stage. 

% - accordingly, we propose AM and EAER. 

% - Reward scores have sharply different scales, which cause instability for PPO training → AM

% - PPO training results in catastrophic forgetting problems on tasks that are well aligned during the SFT stage, which we termed as RL Alignment tax. → Expert Alignments Bonus (EAB)

\section{Preliminary}
\label{sec:background}

In recent machine learning research, RLHF~\citep{ouyang2022training,anthropic} has emerged as a pivotal strategy for aligning LLMs to human goals (e.g. being helpful and harmless).
RLHF typically follows the SFT phase, where SFT aligns a LLM with human objectives using teacher forcing on (prompt, response) pairs.
However, despite this alignment, the LLM may still struggle with generalization when faced with unseen tasks.
% RLHF overcomes this limitation by implementing a reward model (RM) to emulate human feedback through policy learning. 
% In this process, the RM is trained using human preference data, including rankings provided by humans for multiple responses to the same input prompt.

%% Reward modeling
Learning a reward function from interaction between LLMs and humans and optimizing LLMs with the learned reward function using reinforcement learning has been shown as an effective approach to solving the LLM alignment problem. \citealt{leike2018scalable,stiennon2020learning,ouyang2022training} proposed a method involving reinforcement learning from human feedback, where RMs are trained on a dataset of comparisons between two model outputs generated from the same input. The goal is to assign higher rewards to outputs preferred by human labelers over others. Typically, this is achieved by adding a value head that outputs a scalar value on pre-trained transformer-baesd LMs with last umembedding layer removed. Specifically, the reward modeling loss is as follows:
\begin{equation}
\label{eqa:rm_loss}
\begin{aligned}
\mathcal{L}_{\text{RM}} = - E_{(x, y_c, y_r) \sim D^{\mathtt{RM}}} [\log(\sigma(r_\theta(x, y_c) - r_\theta(x, y_r)))] 
\end{aligned}
\end{equation}
where $r_\theta(x,y)$ denotes the reward score for prompt $x$ and response $y$ with parameters $\theta$, $y_c$ is the preferred response of the pair $y_c$ and $y_r$, 
and $D^\mathtt{RM}$ is the complete of comparison dataset.

In what follows, Proximal Policy Optimization (PPO) ~\citep{schulman2017proximal} is commonly adopted as the reinforcement learning algorithm to optimize a policy due to its strengths in stability and simplicity.
Particularly, the PPO objective for policy $\pi$ on a prompt dataset $D$ is defined as:
\begin{equation} \label{eq:ppo}
    \mathcal{L}_{\text{PPO}} = \mathbb{E}_{x \sim D^{\mathtt{PPO}}, y\sim\pi_\phi(x)} \big[ r_\theta(x, y) - \beta \log\big(\pi_\phi(y|x)/\pi^{\mathtt{init}}(y|x)\big) \big]
\end{equation}
% Q: we exclude PTX part in this equation, do we have any evidence (from ourself or other papers) that end-to-edn performance won't change w/ and w/o PTX?
where $r_\theta(x, y)$ represents the reward score on the (prompt, response) pair of $(x,y)$; $\pi^{\mathtt{init}}$ indicates the policy before RLHF, and it is kept constant during RLHF training; $\beta$ is the coefficient for the KL-divergence term.
%which prevents over-optimization towards the RM (a.k.a. reward hacking).

% Q: do we have any exps on RS? if not, we have to skip this part or state some reasons why we don't need to run RS in our exps.
Besides PPO, rejection sampling \citep{llama2} recently gains interests as a simple way for aligning LLMs.
As an offline policy learning algorithm, it adopts an iterative process.
For each iteration $n$, it first constructs a new dataset $D_n$ by selecting $(x,y)$ pairs from the main policy $\pi_\phi$ based on criteria $\mathcal{F}$:
\begin{equation}
    D^{\mathtt{PPO}}_n = \{(x,y) \cdot \mathcal{F}(x,y)| \mathrm{~such~that~} x \sim D^{\mathtt{PPO}}, y \sim\pi_\phi(x)\}
\end{equation}
where a commonly used criteria $\mathcal{F}=\mathbbm{1}_{r_\theta(x,y)\ge\tau}$  includes only the samples with RM scores exceed a certain threshold $\tau$.
The policy is then updated by teacher forcing on $D_n^{\mathtt{PPO}}$:
\begin{equation}
    \mathcal{L}_{\text{RS}} = \mathbb{E}_{(x,y)\sim D^{\mathtt{PPO}}_n} \sum_{t=1}^{|y|} \pi_\phi(y_t|y_{<t},x)
\end{equation}

\section{Approach}
\label{sec:approach}

% - Advantage Model
%     - AM score = RM(x, y ) - $\sum_y p(y) * RM (x,y)$ to enforce all the examples share the same measuring scale.
    
% - Expert Alignments Bonus

%     - Prevent model from descending too far off the already aligned SFT model. The distance is measured by KL distances on experts aligned examples.
%     - Expert examples selection
%         - Embedding
%         - AM/RM Score
%         - Entropy
%         - Confidence Score
        
%     - Approaches to integrating to PPO
%         - expert aligned examples reward bonus: $AM score - kl_coef * kl - kl_coef_expert * kl_expert_examples$
%         - general entropy bonus
%         - FTX

\subsection{From Reward Model to Advantage Model}

 The learning objective of equation~\ref{eqa:rm_loss} primarily allows models to distinguish between human-preferred responses and alternative options. It relies only on score differences to assess the likelihood of one response being superior to another. In such case, two different model responses that are both preferred by humans could have dramatically different values. In addition, interpreting the scalar values themselves can be challenging. 

In light of these considerations, we introduce the Advantage Model (AM) for reward modeling. Analogous to the concept of the advantage function in reinforcement learning, the Advantage Model, denoted as $a(x,y)$, quantifies the additional reward that response $y$ can achieve over the expected reward $e$ for prompt $x$. This is formally defined as:
\begin{equation} \label{eq:am}
\begin{aligned}
% a_\theta(x,y) = r_\theta(x,y) - \mathbb{E}_{(x,y)\sim \pi^{\text{init}}(x)}[r_\theta(x,y)] 
a_\theta(x,y) = r_\theta(x,y) - \mathbb{E}_{y\sim\pi^\prime(x)}[\frac{\pi_\phi(y|x)}{\pi^\prime(y|x)}  r_\theta(x,y)] 
\end{aligned}
\end{equation}
Here, the notation $y\sim \pi^\prime(x)$ signifies all possible responses generated by a policy $\pi^\prime(x)$ when given the input prompt $x$. Since the comparison data is typically collected in many batches with different SFT or PPO models, we introduce $\frac{\pi^\phi(y|x)}{\pi^\prime(y|x)}$, the importance weight term to negate the bias introduced by the policy distribution shift. Intuitively, the extra reward gains of good response $y_c$ and the reward losses of bad response $y_r$ should be bounded by a margin $m$. As such, the training objective of AM consists of two parts, \textit{ranking loss} that aligns with the formulation in Equation \ref{eqa:rm_loss}, and \textit{bounding loss} to ensure the well-calibrated bounding of AM scores. It is formally defined as follows:
\begin{equation}
\begin{aligned}
\mathcal{L}_{\text{AM}} = - E_{(x, y_c, y_r) \sim D^\mathtt{RM}} [\log(\sigma(a_\theta(x, y_c) - a_\theta(x, y_r))) \\  +~\log(\sigma(m(x) - a_\theta(x, y_c))) +~\log(\sigma(m(x) + a_\theta(x, y_r)))]
\end{aligned}
\end{equation}
where $m(x)$\footnote{We think that $m(x)$ may have a connection with the complexity or difficulty involved in learning the reward function for prompts similar to $x$. However, this is speculative and requires further investigation. We leave this aspect as a topic for future study and exploration. Throughout our experiments, we set $m(x)$ as 2.5.} is the function that defines the permitted margin for prompt $x$. However, it is infeasible to list every potential response to calculate the expected reward. To address this, we propose parameterizing the expected reward of the current policy, denoted as:
\begin{equation}
\begin{aligned}
e_\tau(x) = \mathbb{E}_{y \sim \pi_\phi(x)}[r_\theta(x,y)]
\end{aligned}
\end{equation}

By integrating the term representing the importance weight, we can reformulate the equation ~\label{eq:am}  as follows:
\begin{equation} \label{eq:am2}
\begin{aligned}
a_\theta(x,y) = r_\theta(x,y) -  \tfrac{N - K}{N} e_\tau(x) - \sum_{k=1}^K \tfrac{1}{N} \tfrac{\pi^\phi(y|x)}{\pi^\prime_k(y|x)} r_\theta(x,y)
\end{aligned}
\end{equation}
where $N$ serves as a hyperparameter that harmonizes the emphasis placed on the current policy model relative to alternate policy models. $K$ specifies the number of alternate policy models utilized for comparison data collection. Additionally, $\pi^\prime_k(y|x)$ indicates the probability derived from the $k$th policy model.

\subsection{PPO with Selective Rehearsal}

In addition, we propose Selective Rehearsal to maintain the skills that are already acquired before RLHF.
Selective rehearsal takes two major steps: representative example discovery and rehearsal training.

\paragraph{Representative example discovery}
Given the policy $\pi_\phi$ and PPO training prompts with policy outputs $D^\mathtt{PPO} = [(x_1, y_1), (x_2, y_2) \dots]$,
our goal is to select high-quality $(x,y)$ pairs from $D^\mathtt{PPO}$ that cover as many skills (e.g., solving algebra problems and writing resume) as possible.
In order to let selected $(x,y)$ pairs represent as many skills as possible, we first adopt a clustering algorithm (e.g. KMeans or Gaussian mixture) to separate $D^\mathtt{PPO}$ into $c$ clusters.
To assure the representativeness and quality of the selected data, we only keep certain $(x,y)$ pairs within each cluster that satisfy certain criteria regarding aspects such as advantage (reward) model score, entropy (low entropy indicates high confidence), human satisfaction rate or response length (higher length may indicate redundancy).

Here we adopt the SimCSE \citep{gao2021simcse} sentence embedding\footnote{\url{https://huggingface.co/princeton-nlp/sup-simcse-roberta-base}} to represent the query $x$ for each $(x,y)$ pair before running a KMeans algorithm on these embeddings to be grouped into $c$ clusters.
We briefly study the influence of cluster number $c$ in Section \ref{sec:eval}.
Within each cluster, here we simply choose the top-$k$ $(x,y)$ pairs with the highest advantage model score (Eq. \ref{eq:am}).
We leave other strategies (e.g. combining advantage score with entropy score) in future work.

One reason we select our rehearsal data from the PPO training data with each response $y$ being generated from the initial policy model is to enable a more fair and nuanced comparison, as no additional information is introduced.
In other scenarios, the rehearsal $(x,y)$ pairs could come from other important data sources representing specific skills (e.g. math-problem solving) the main policy are not expected to forget.

\paragraph{Rehearsal training}
After obtaining the rehearsal $(x,y)$ pairs of all clusters, we shuffle them together to form the rehearsal dataset $D_R$ and compute NLL loss on $D_R$ as a supplement to the standard PPO loss defined in Equation \ref{eq:ppo}:
\begin{equation}
    \mathcal{L}_{\text{PPO-SR}} = \mathcal{L}_{\text{PPO}} + \gamma \mathbb{E}_{(x,y)\sim D_R} \sum_{t=1}^{|y|} \pi_\phi(y_t|y_{<t},x)
\end{equation}
where the coefficient for the NLL loss $\gamma$ is empirically set to $0.01$.

Rehearsal training is similar with rejection sampling and reinforced self-training \citep{gulcehre2023reinforced} by using self-generated $y$s of high reward model score for supervised training.
However, rehearsal training captures multi-dimensional important aspects (e.g., diversity), while rejection sampling and reinforced self-training only consider reward model score.

Alternatively, one can view selective rehearsal as a means of amplifying the weight of the KL-divergence term in PPO training (Eq. \ref{eq:ppo}) for crucial instances and their related counterparts.

\section{Experiments}
\label{sec:experiments}
% \subsection{Advantage Model}
\subsection{Datasets and Models}

\paragraph{RM datasets}
We conducted experiments on both English and Chinese datasets. For the English experiments, we utilized the HH-RLFH dataset \citep{anthropic,ganguli2022red},
which comprises 118k helpful and 42k harmless examples for training, and 8.5k for testing. It is worth noting that many studies train different RMs separately for helpful and harmless examples to achieve better performance. However, in our experiments, we did not distinguish between helpful and harmless examples.

For the Chinese dataset, we collected comparison examples with quantities similar to those used in LLaMA 2~\citep{llama2}.
Our annotation procedure operates as follows: First, we ask annotators to generate prompts based on a task spectrum. Next, we sample five responses from the same SFT model using varied sampling hyper-parameters. Finally, we distribute these responses to five annotators for ranking based on provided criteria. Following~\cite{anthropic}, the annotation criteria focuses on helpfulness and harmless. %For the expert-aligned testset, 

\paragraph{PPO dataset}
We sampled queries from two popular domain-general datasts, COIG\footnote{\url{https://huggingface.co/datasets/BAAI/COIG}} and firefly\footnote{\url{https://huggingface.co/datasets/YeungNLP/firefly-train-1.1M}} to form our PPO dataset.
Particularly, we obtained 64,364 and 2,623 for PPO training and testing, respectively\footnote{The PPO training and testing query sets could be shared upon request.}.
There is no intersection between the training and testing sets. Additionally, we selected 1,704 examples from the SFT test data to create a \textit{forget test set}, enabling us to evaluate the model's ability to retain learned knowledge.

\paragraph{Models}
We employed BLOOMZ \citep{muennighoff2022crosslingual} as our pre-trained model backbone. More specifically, BLOOMZ$_{\mathtt{7B}}$ was used for reward modeling and BLOOMZ$_{\mathtt{176B}}$ was used for SFT and RLHF training. 

\subsection{Training Setups}

We initialized our models using pre-trained checkpoints. The architectural configuration and hyper-parameters were kept consistent with those of the pre-trained models, except that a value head is added to produce a scalar reward. A learning rate of 5e-6 was employed, coupled with a warm-up strategy covering the initial 10\% of training steps and a cosine learning rate schedule decreasing to 10\% of the initial learning rate. For the English dataset, a global batch size of 180 was employed, whereas for the Chinese dataset, the batch size was set to 480. The Overfitting issue is observed in general after models are trained for one epoch. As such, we fixed the training epoch as 1 for the all the experiments.For PPO training, a learning rate of $5\times 10^{-7}$ and a global batch size of 256 is employed.
The actor model is trained for 100 steps for all experiments. The SFT model is trained on the proprietary dataset. We omit these details since these are not the focus of this paper.

\subsection{Evaluation} \label{sec:eval}
\begin{table*}%
	\centering
	\setlength{\tabcolsep}{1.5mm}{
		\begin{tabular}{lcccc}
			\toprule
			\multirow{2}{*}{Model} & \multicolumn{2}{c}{HH-RLHF} & \multicolumn{2}{c}{Proprietary Data} \\
			\cmidrule(l){2-3} \cmidrule(l){4-5}
                & $\mathtt{Accuracy}$ $\uparrow$ & $\mathtt{ECE}$ $\downarrow$ & $\mathtt{Accuracy}$ $\uparrow$ & $\mathtt{ECE}$ $\downarrow$\\
            \midrule
            OpenAssistant~\cite{kopf2023openassistant} & 69.24 & - & - & - \\
            \midrule
            Reward Model & 69.25 & 4.70 & 74.75 & 5.35 \\
            Advantage Model & 69.43 & 3.48 & 75.28 & 3.83 \\

			\bottomrule
		\end{tabular}
	}
		    
	\caption{Evaluation results on HH-RLHF and our proprietary data. Note that maximizing accuracy is not the exclusive objective in AM optimization. The aim also extends to reducing ECE to improve reliability, whilst sustaining or improving the level of ranking accuracy compared with RM. }
	\label{tab:reward_model}
\end{table*}
\paragraph{AM Evaluation Results} 

Firstly, we present the overall accuracy and Expected Calibration Error (ECE) for both RM and AM on each dataset. For the English dataset, we additionally compare our method with the publicly available OpenAssistant~\citep{kopf2023openassistant} which utilized DeBERTa~\citep{he2020deberta} for reward modeling. Table~\ref{tab:reward_model} lists all the results. We observe that AM achieves slightly higher accuracy but significantly lower ECE on all the datasets. This indicates that AM is capable of maintaining the same level of ranking accuracy while providing reliable and well-calibrated scores. A detailed analysis of calibrations is provided in the following sections. We attribute this phenomenon to the fact that AM is formulated to directly model additional rewards, \ie advantages, making it more stable and less prone to yield high variances cores. Additionally, the accuracy on the proprietary data is much higher than that on HH-RLHF. We speculate that the trade-off between helpfulness and harmlessness objectives is more pronounced in HH-RLHF, possibly due to the limited presence of harmful examples in our proprietary data.

\paragraph{Calibrations of AM} 
\begin{figure}[htbp]
    \centering
    \begin{subfigure}[b]{0.46\textwidth}
        \centering
        \includegraphics[width=\textwidth]{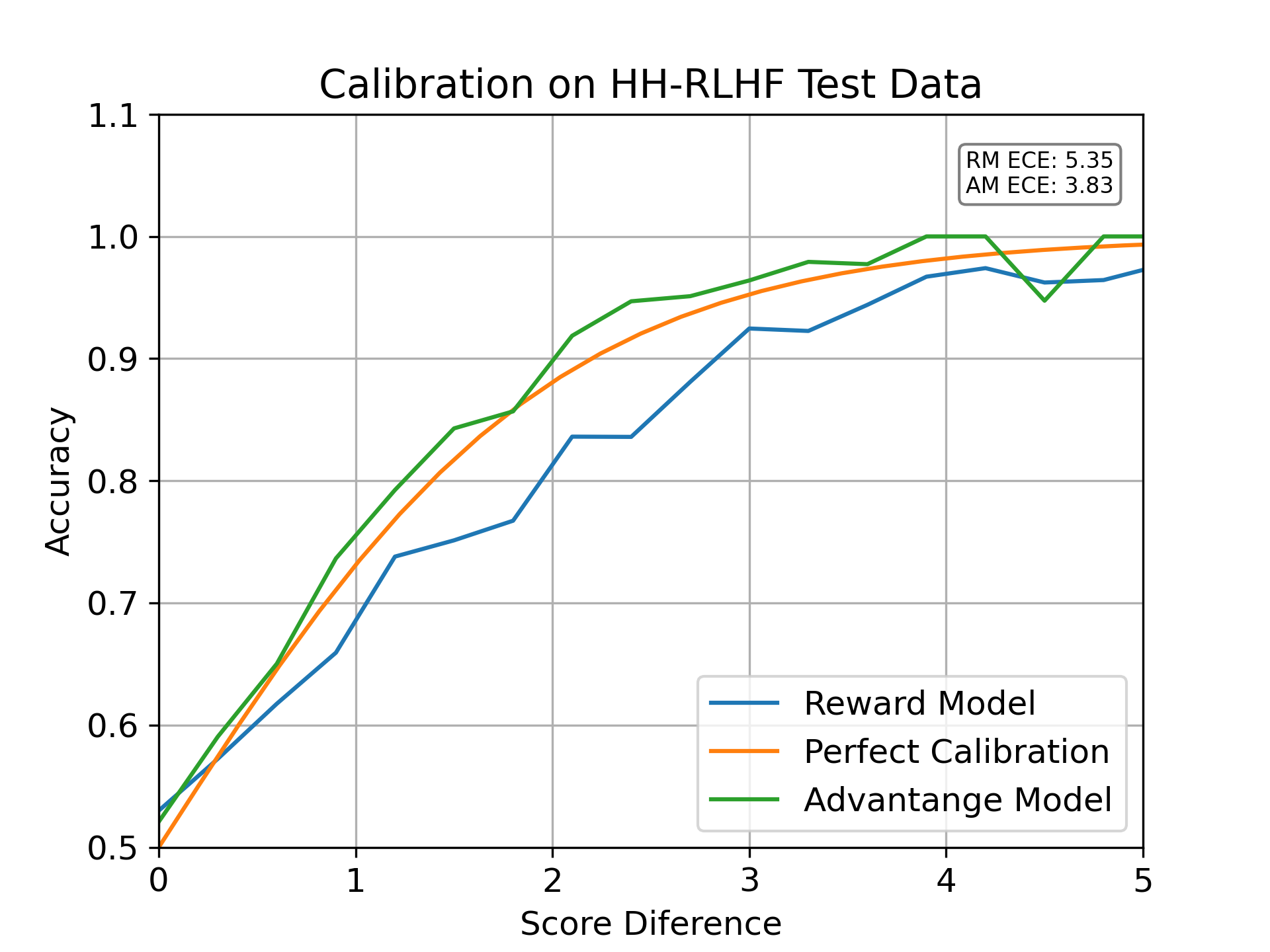}
        % \caption{A}
        \label{fig:A}
    \end{subfigure}
    \quad
    \begin{subfigure}[b]{0.46\textwidth}
        \centering
        \includegraphics[width=\textwidth]{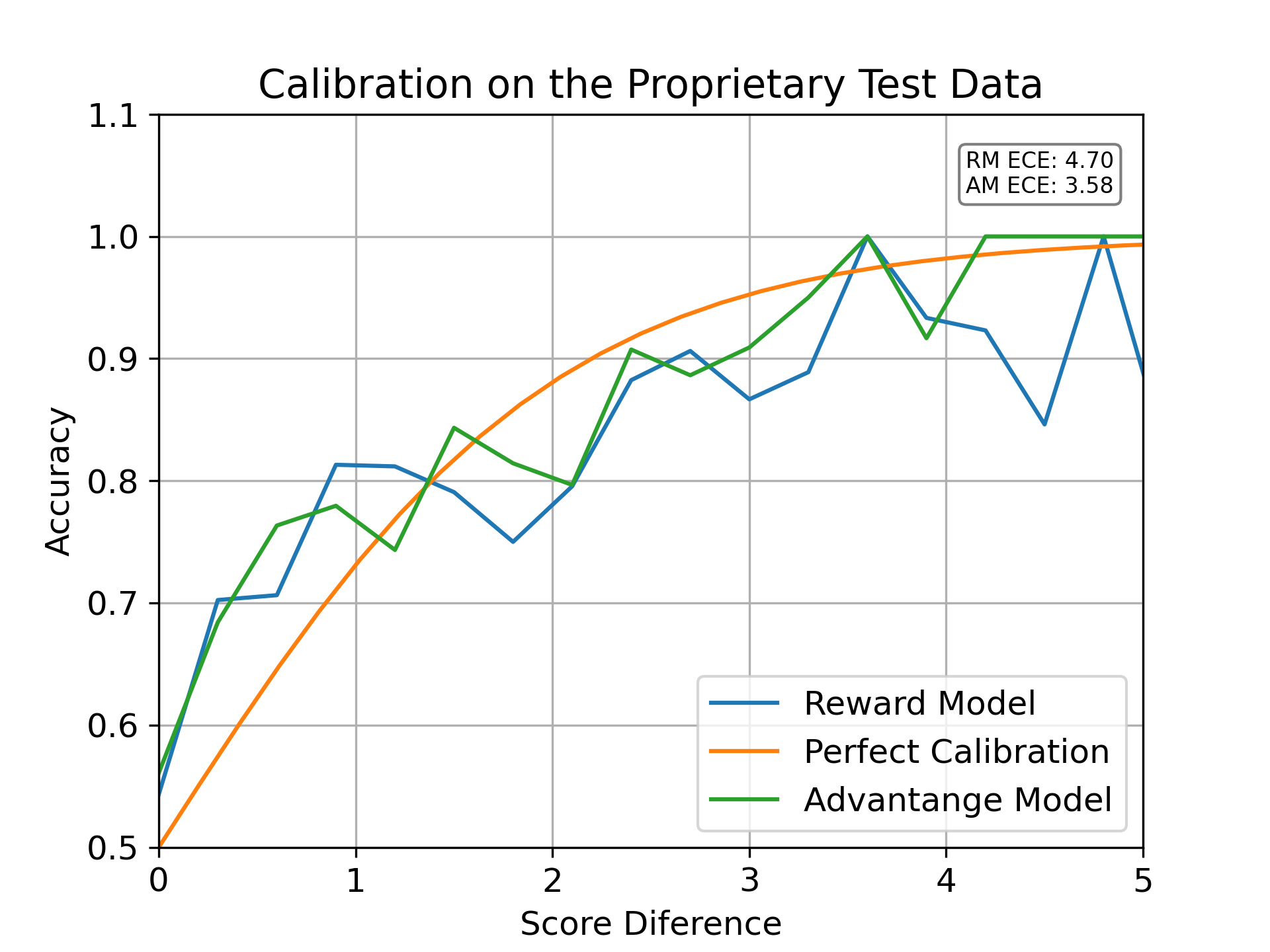}
        % \caption{B}
        \label{fig:B}
    \end{subfigure}
    \caption{Ranking accuracy is shown as a function of the difference in scores between higher and lower ranked responses. The orange lines indicate the calibrated prediction of accuracy $1 / (1 + e^{-\Delta})$ in which $\Delta$ denotes the score difference. On the left, we show calibration of RM and AM on HH-RLHF data while on the right we show results for our proprietary data. We observe that AM calibration is better than RM's.}
    \label{fig:calibration}
\end{figure}
\begin{figure}[htbp]
    \centering
    \begin{subfigure}[b]{0.46\textwidth}
        \centering
        \includegraphics[width=\textwidth]{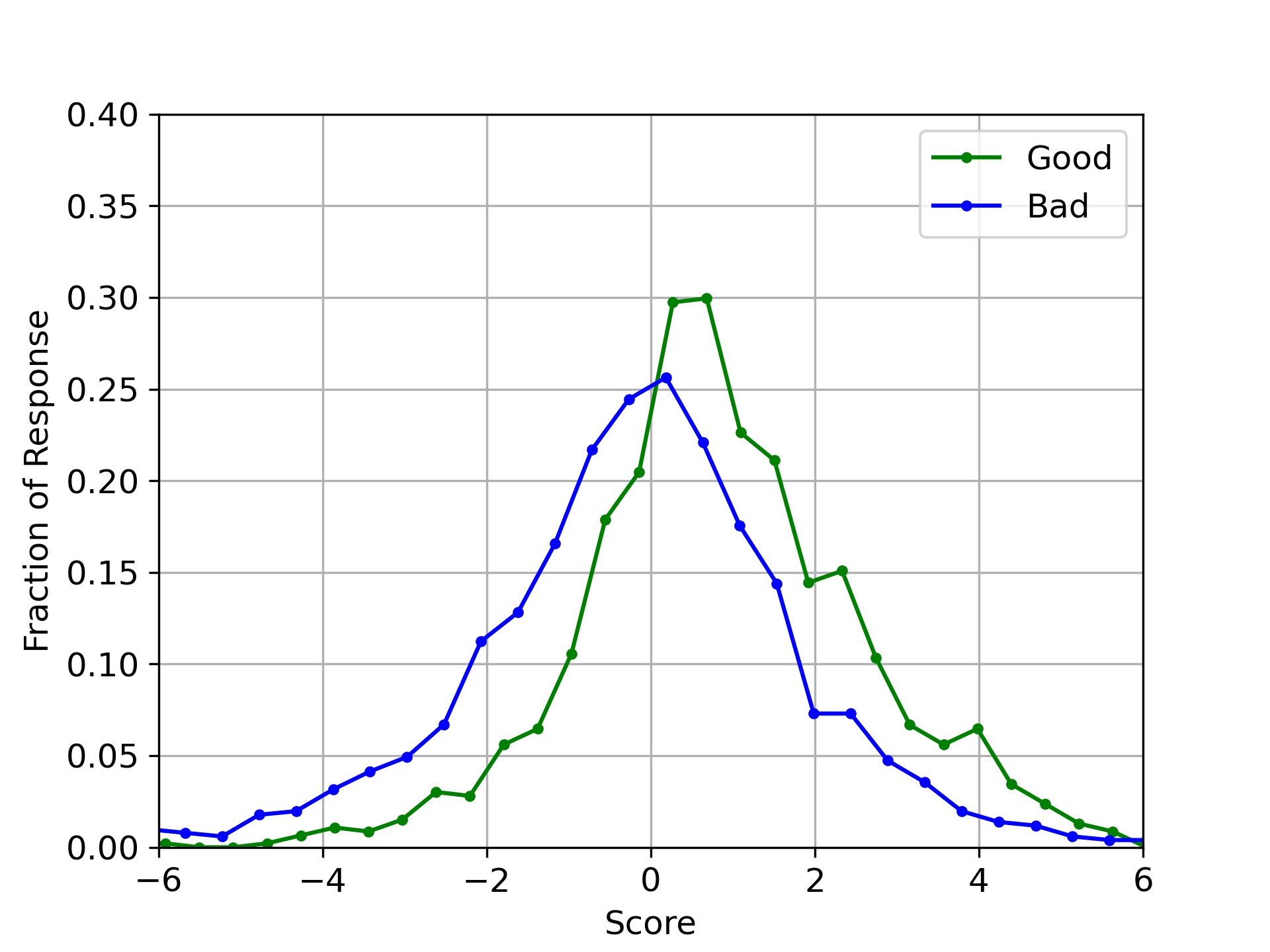}
        \caption{RM score distribution.}
        \label{fig:A}
    \end{subfigure}
    \quad
    \begin{subfigure}[b]{0.46\textwidth}
        \centering
        \includegraphics[width=\textwidth]{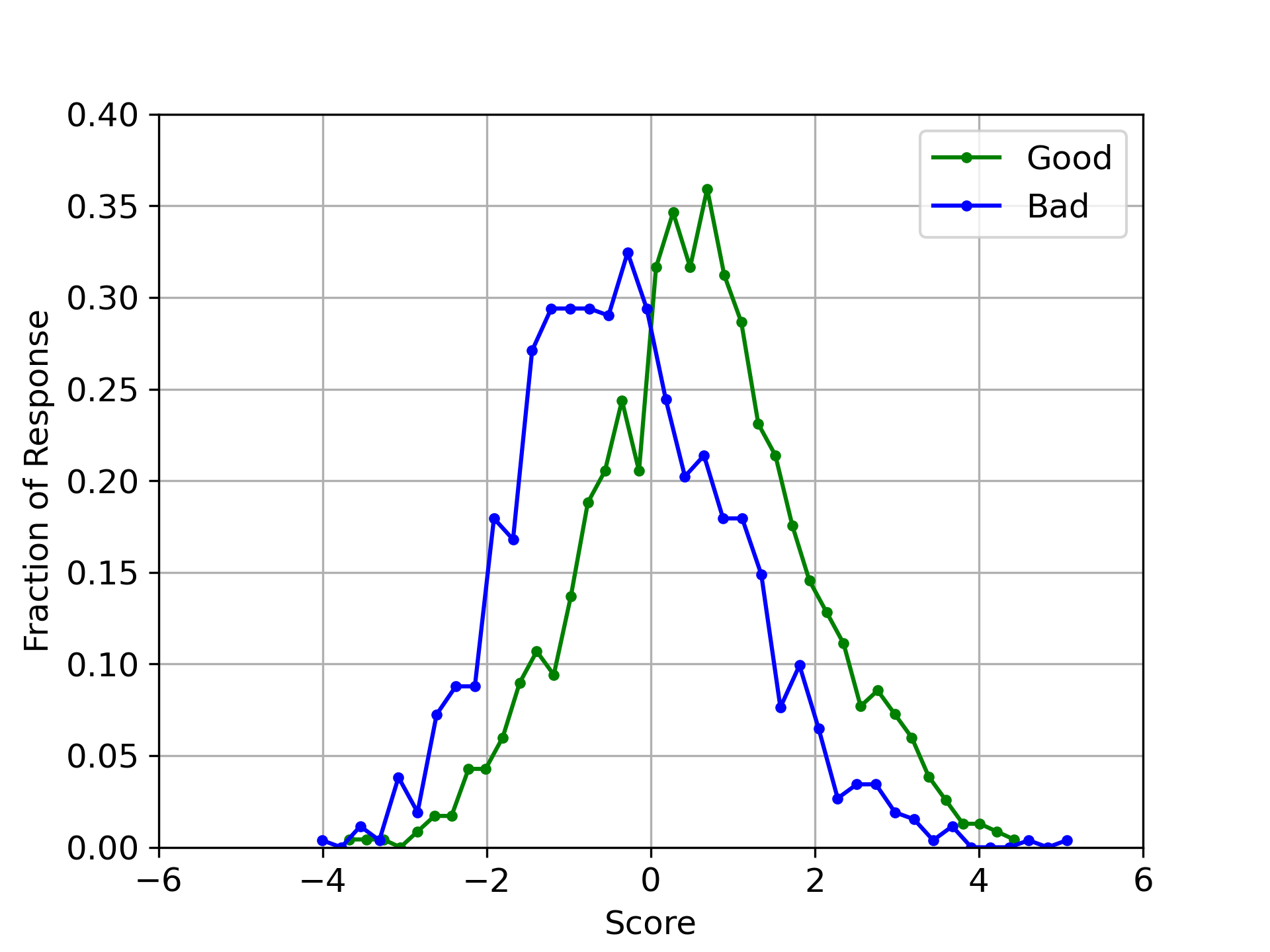}
        \caption{AM score distribution.}
        \label{fig:B}
    \end{subfigure}
    \caption{Distributions of RM and AM scores for pairs of good and bad examples from the proprietary data.}
    \label{fig:score_distribution}
\end{figure}
\vspace{-2mm}
The reward model score of a response should accurately reflect the probability that humans prefer it. These probabilities must be precise; in other words, the scores should be well-calibrated. This is crucial since these scores will serve as reward signals to guide PPO training~\cite{anthropic}. To assess whether our AM is calibrated or not, in Figure~\ref{fig:calibration}, we depict the ranking accuracy as a function of score differences assigned to pairs of samples. An orange line representing perfect calibration is also included. Our observations indicate that the AM exhibits significantly lower ECE and is better calibrated than RM on both datasets, whereas RM tends to be overconfident in most cases. We further show the distribution of scores for both good and bad examples in Figure~\ref{fig:score_distribution}. While in general both RM and AM are able to assign higher scores for good examples, AM exhibits a more distinct distribution pattern.

\paragraph{Means and variances of AM}
\begin{figure}[htbp]
    \centering
    \begin{subfigure}[b]{0.46\textwidth}
        \centering
        \includegraphics[width=\textwidth]{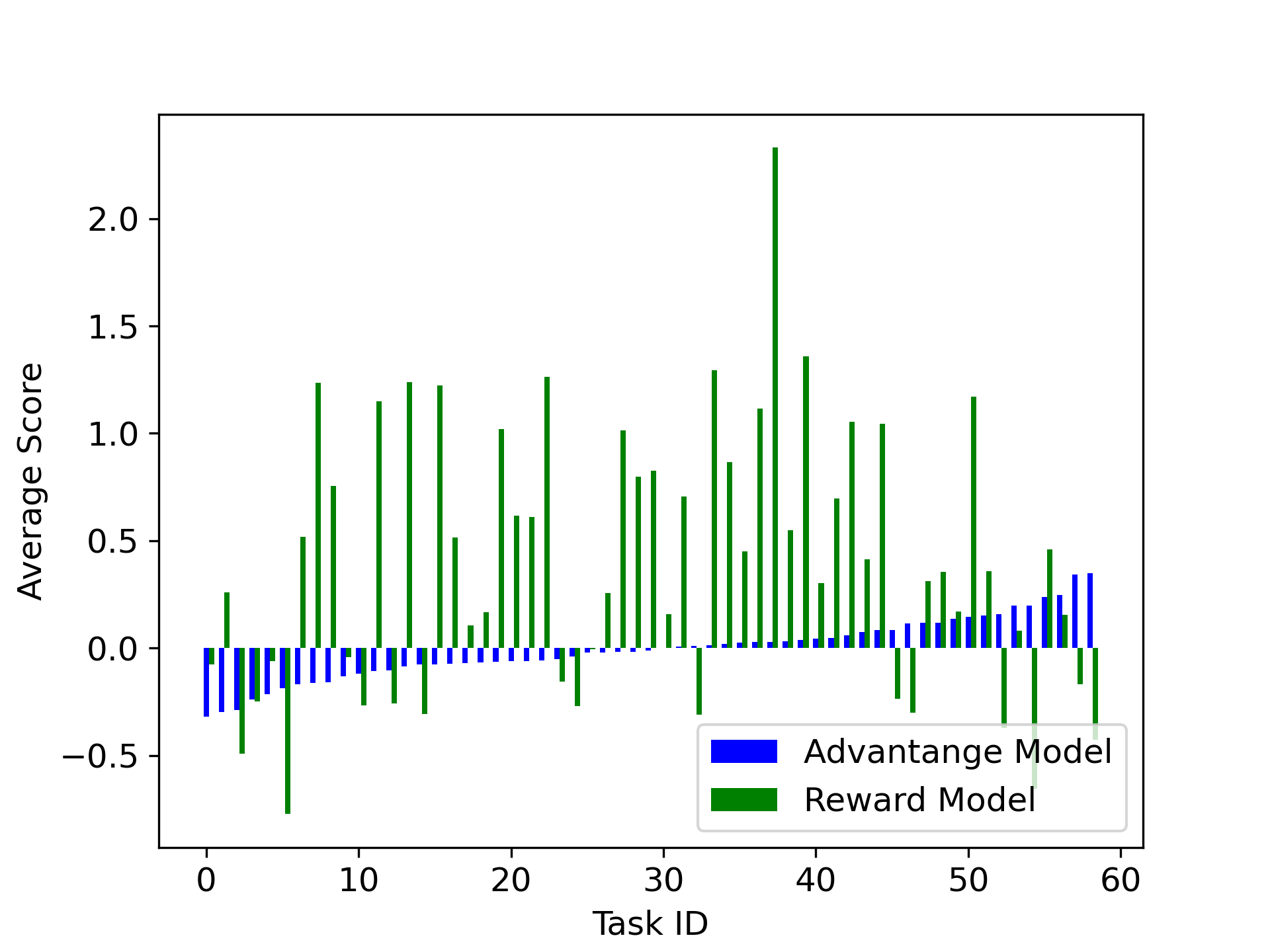}
        \caption{Mean scores of RM and AM for each task.}
        \label{fig:am_mean}
    \end{subfigure}
    \quad
    \begin{subfigure}[b]{0.46\textwidth}
        \centering
        \includegraphics[width=\textwidth]{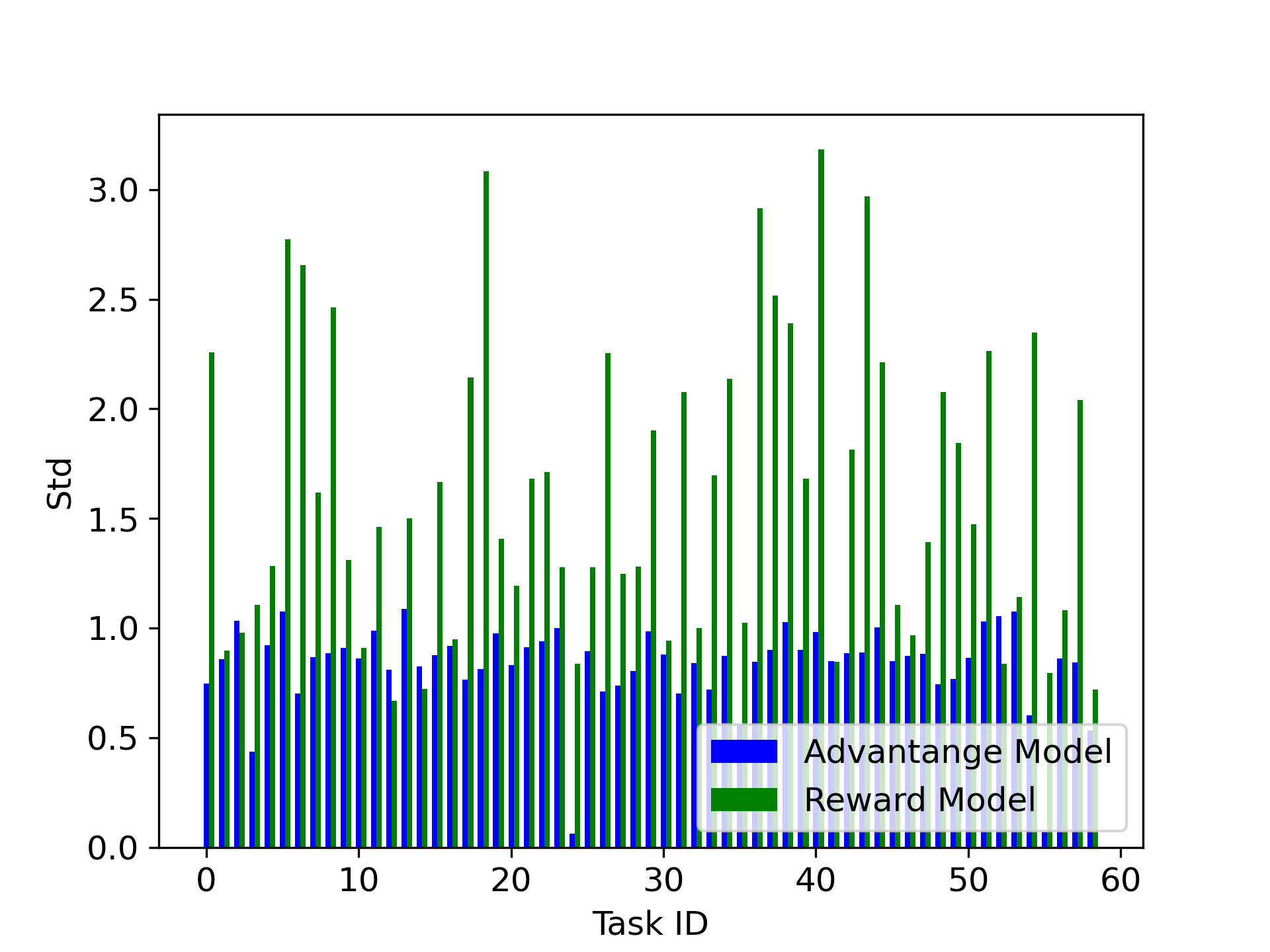}
        \caption{Std of RM and AM for each task.}
        \label{fig:am_std}
    \end{subfigure}
    \caption{Mean and standard variance for each task categorized by a task spectrum on the in-house data.}
    \label{fig:am_mean_std}
\end{figure}

During PPO training, RLHF exhibits instability, largely owing to unpredictable fluctuations in reward estimation scales. Directly modeling advantage, as our AM does, could potentially alleviate the above issue. To validate AM's efficacy in stabilizing score scales and ranges, we calculated the AM scores for individual examples and analyzed the mean and variance across all the the task spectrum. This analysis is depicted in Figure~\ref{fig:am_mean}. We observe markedly different means for each task in the case of RM. Such significant disparities in means can potentially give rise to reward hacking issues~\citep{skalse2022defining} and result in repeated failures during PPO training. In addition, Figure~\ref{fig:am_std} illustrates the standard deviations of both AM and RM, with AM consistently operating at a stable scale. These results endorse AM as a strategy designed to normalize reward scores at the individual example level while enhancing ranking accuracy.

% It is evident from this figure that AM consistently exhibits variance at a stable scale. Given these observations, AM could be viewed as an approach aimed at normalizing reward scores at the example level, while simultaneously optimizing it to ensure ranking precision.

\paragraph{PPO training results}
\begin{figure}[htbp]
    \centering
    \begin{subfigure}[b]{0.46\textwidth}
        \centering
        \includegraphics[width=\textwidth]{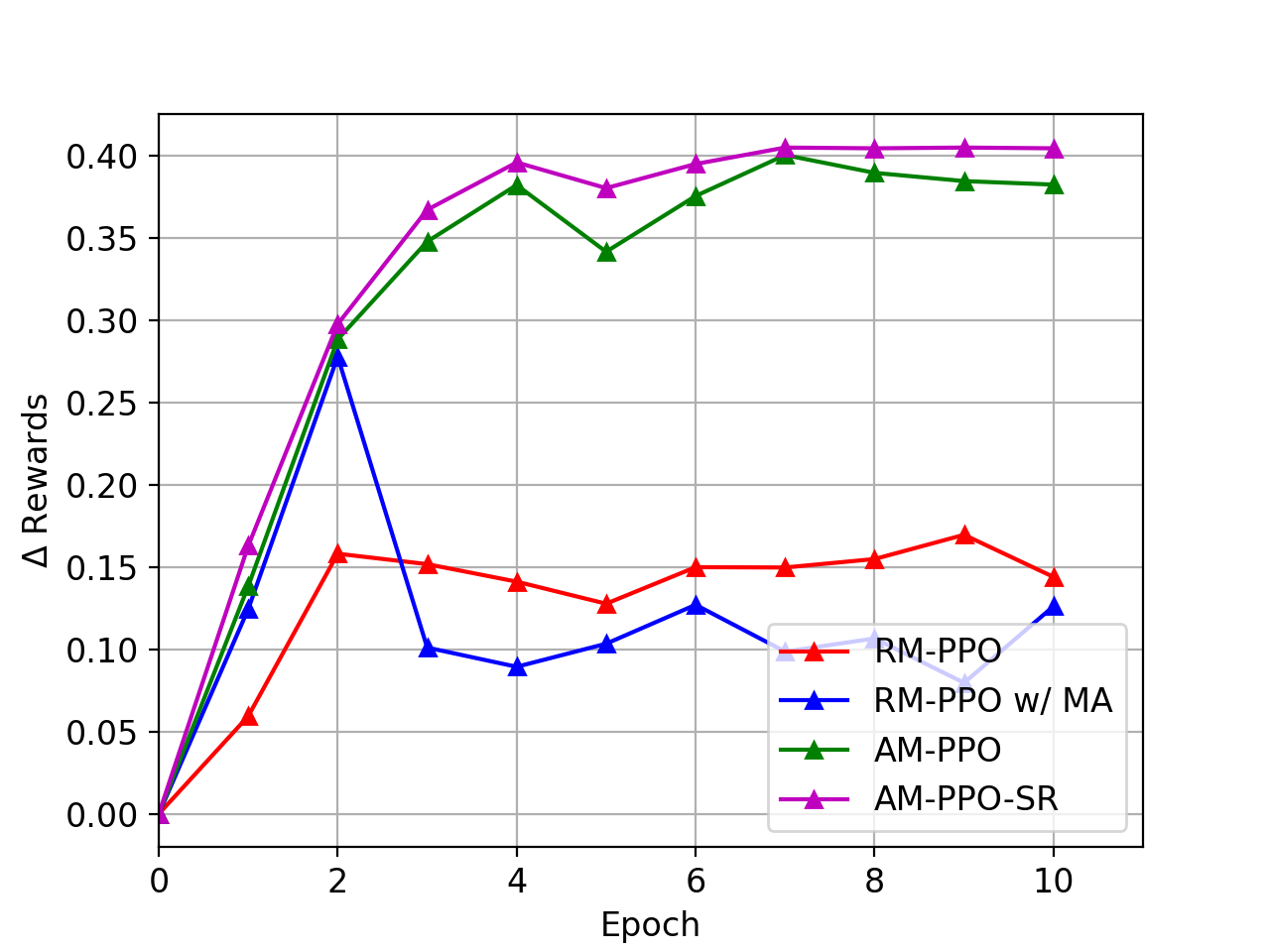}
        \caption{Learning curves of various models on delta rewards}
        \label{fig:ppo_learning_curve_reward_reward}
    \end{subfigure}
    \quad
    \begin{subfigure}[b]{0.46\textwidth}
        \centering
        \includegraphics[width=\textwidth]{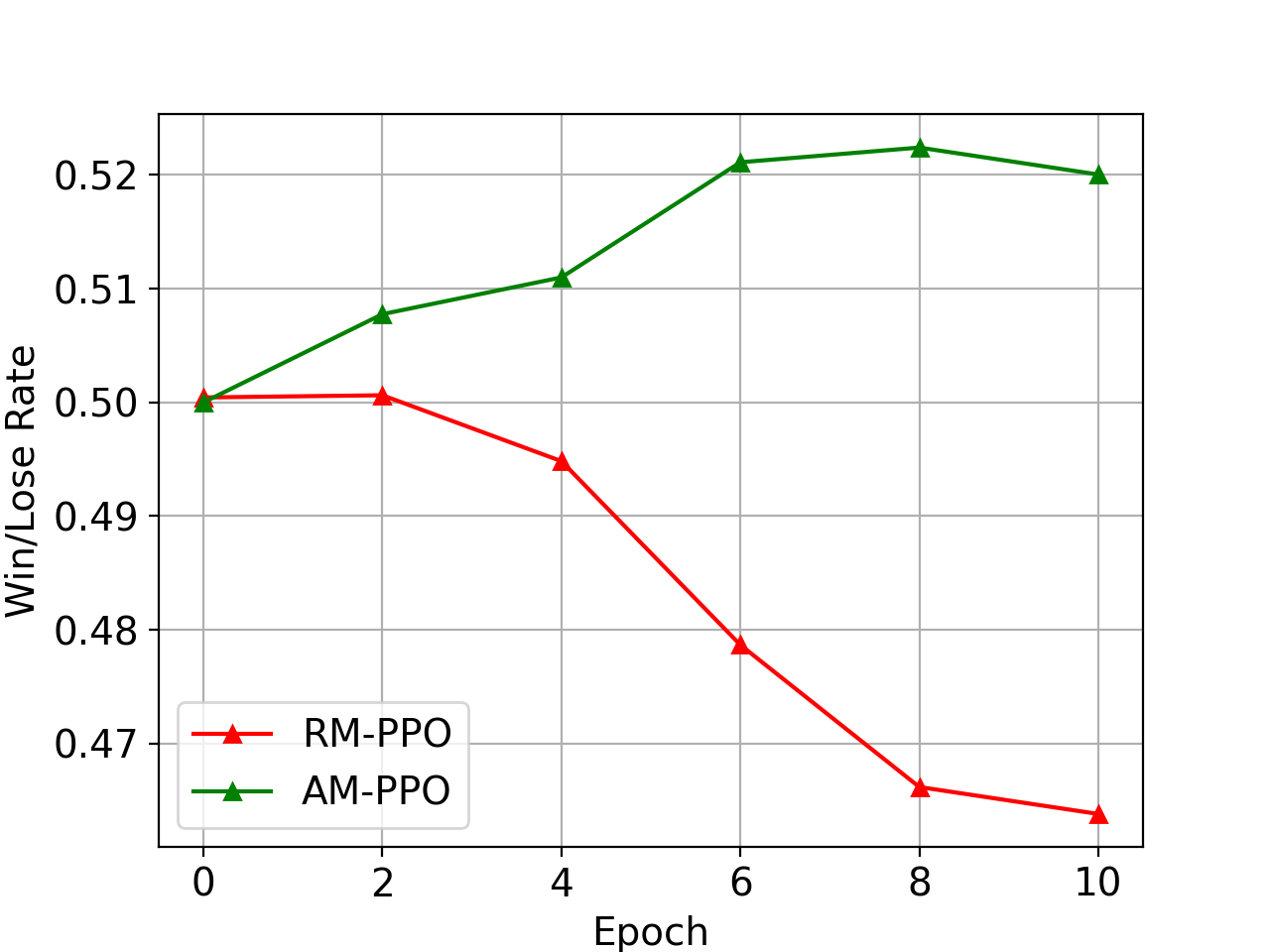}
        \caption{Win/Loss rate over SFT model evaluated by GPT-4.}
        \label{fig:gpt4_ppo_learning_curve}
    \end{subfigure}
    \caption{PPO training curves on the Main Test Set with different scoring models. RM-PPO and AM-PPO denote PPO trained with Reward Model and Advantage Model, respectively. AM-PPO-SER additionally equips with Selective Rehearsal.}
    \label{fig:ppo_learning_curve}
\end{figure}

% \begin{wrapfigure}{r}{width=0.5\linewidth}
%   \includegraphics[width=0.5\linewidth]{figures/ppo_learning_curve.png}
%   \caption{Figure caption}
% \end{wrapfigure}

% \begin{wrapfigure}{r}{8cm}
%   \includegraphics[width=8cm]{figures/ppo_learning_curve.png}
%   \caption{Your caption here}
% \end{wrapfigure}
% Figure~\ref{fig:ppo_learning_curve} shows the PPO training curve. Firstly, we observe that the AM-PPO method consistently outperforms the RM-PPO with doubled reward gains.
We conducted a comparative analysis of PPO training with different scoring models in terms of their performance on both main test set and forget test set. The learning curve is shown in~\ref{fig:ppo_learning_curve}. We observe that AM-PPO outperformed RM-PPO in the main set, achieving higher rewards and a superior win rate over the SFT model. In addition, RM-PPO faces significant reward hacking issues, witnessed by a drop in win rate evaluated by GPT-4, shown in~\ref{fig:gpt4_ppo_learning_curve} despite a rise in RM scores. Despite utilizing moving average for score normalization, RM-PPO w/ MA encounters instabilities during PPO training.  Conversely, AM-PPO exhibits resistance to such problems, maintaining stable GPT-4 outcomes. This emphasizes AM's stability and alignment efficiency over RM. The forget test set result reveal RM-PPO's substantial susceptibility to catastrophic forgetting, portraying a noticeable performance drop. In contrast, AM-PPO is stable, avoiding significant drops and showcasing stability. Incorporating selective rehearsal, the AM-PPO-SR variant demonstrate an uplifted win rate on both sets, underscoring the role of selective rehearsal in alleviating catastrophic forgetting and enhancing model efficacy.

% main results
\begin{table*}%
	\centering
	\setlength{\tabcolsep}{1.5mm}{
		\begin{tabular}{lcccccc}
			\toprule

   \multirow{2}{*}{Model} & \multicolumn{3}{c}{Main Test Set} & \multicolumn{3}{c}{Forget Test Set} \\
   \cmidrule{2-4} \cmidrule{5-7}
			& $\mathtt{Win}$ $\uparrow$  & $\mathtt{Lose}$ $\downarrow$ & Tie & $\mathtt{Win}$ $\uparrow$ & $\mathtt{Lose}$ $\downarrow$ & Tie \\
            \midrule
            
            RM-PPO  & 12.72 & 12.62 & 74.66 & 16.87 & 29.28 & 53.84 \\
            AM-PPO  & 14.87 & 10.38 & 74.74 & 9.70 & 8.44 & 81.86 \\
            AM-PPO-SR  & 15.78 & 9.77 & 74.45 & 10.30 & 7.95 & 81.75 \\
			\bottomrule
		\end{tabular}
	}
		    
	\caption{Comparison results of different models over the SFT model.}
	\label{tab:reward_model}
\end{table*}

\begin{wrapfigure}{L}{0.55\textwidth}
    \centering
    \includegraphics[width=0.5\textwidth]{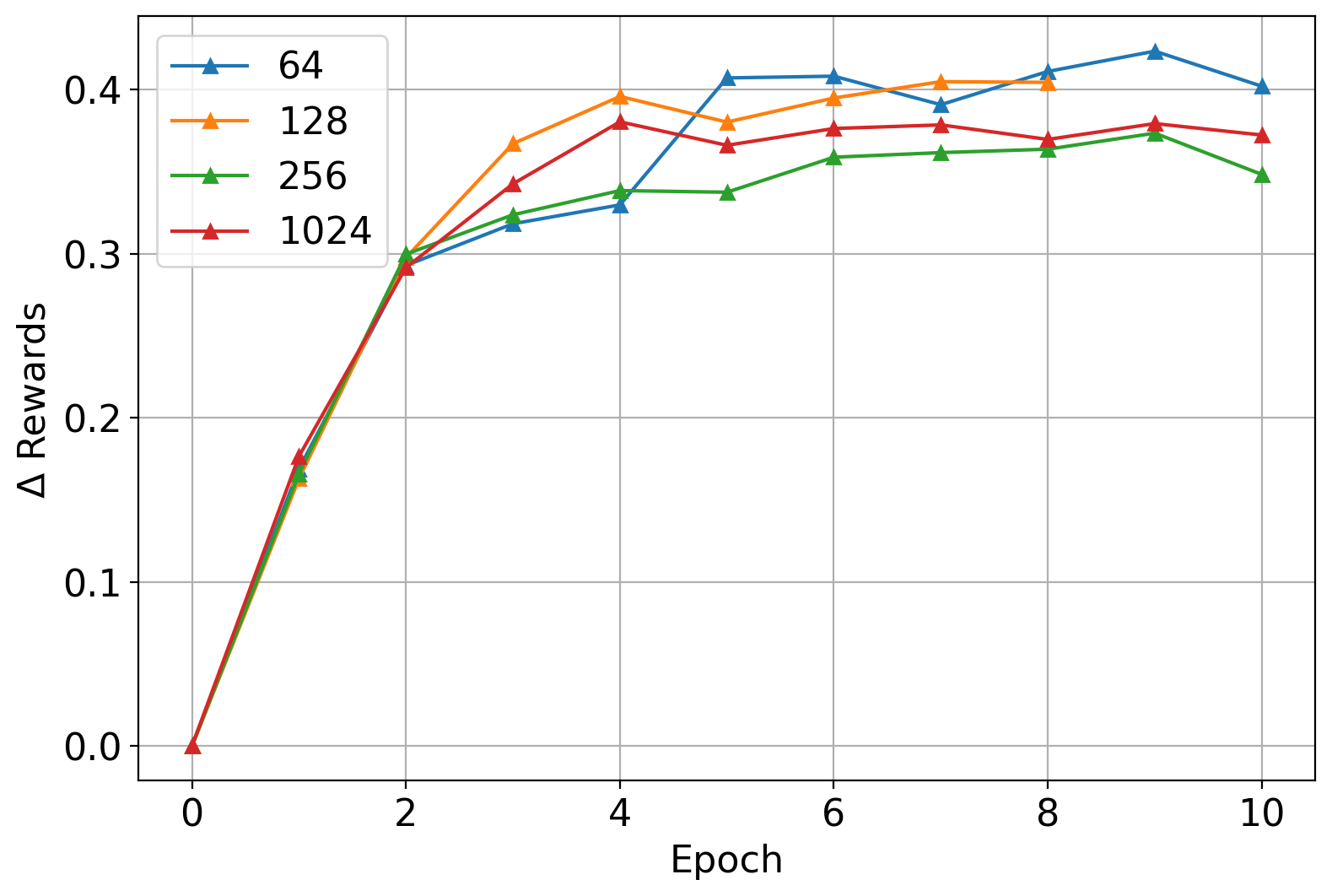}
\caption{The AM-PPO-SR training curves on the Main Test Set with different number of clustering groups $c$ for selective rehearsal.}
\label{fig:rehearsal}\vspace{-9mm}
\end{wrapfigure}

\paragraph{Analysis on Selective Rehearsal}
We also conduct an in-depth examination of the impact of the number of clusters, denoted as $c$, in the context of selective rehearsal during PPO training.
As illustrated in Figure \ref{fig:rehearsal}, our results reveal a relatively consistent variance of approximately 0.05 points in test-set rewards across various cluster numbers $c$.
While our findings highlight the robustness of the selective rehearsal technique, we recommend conducting a thorough analysis of this aspect when applying selective rehearsal to different datasets, as domain-specific variations can have a notable impact.
\\
\section{Related Work}
\label{sec:related_work}

\paragraph{LLM Alignments with Human Preferences.} LLMs are typically pre-trained on extensive datasets and can be adapted to a wide variety of downstream tasks. One critical aspect of utilizing LLMs effectively is ensuring their alignment with human preferences, which helps in averting responses that are unsafe, toxic, sexually explicit, biased, or criminal~\citep{leike2018scalable}. A predominant strategy in achieving this is RLHF. This involves training a reward model based on human feedback and utilizing PPO to improve to fine-tuning LLMs~\citep{christiano2017deep, anthropic, glaese2022improving, bai2022constitutional, stiennon2020learning, qiu2022valuenet}. 

\paragraph{Instabilities in RLHF.} Despite its success, the RLHF approach is inherently complex and poses significant challenges, thereby encouraging the exploration of simpler methods to align LLMs with human preferences. In this context, \citet{cobbe2021training} introduced the best-of-n sampling, which reinforces LLMs by choosing the responses with the highest reward score from a set of n responses. A similar pathway was pursued by RAFT~\citep{dong2023raft}, which focuses on selecting high-quality samples to fine-tuning to enhance the model's performance. Moreover, the RRHF strategy~\citep{yuan2023rrhf} evaluates sampled responses from various sources using the logarithm of conditional probabilities. It then aligns these probabilities with human preferences by applying ranking loss, fostering a more refined alignment process.  Furthermore, \citet{rafailov2023direct} introduced the concept of Direct Preference Optimization (DPO). This approach leverages a relationship between reward functions and optimal policies to address a constrained reward maximization problem through a single stage of policy training. In a similar vein, Preference Ranking Optimization (PRO)~\citep{song2023preference} sidesteps the necessity for Reinforcement Learning (RL) training. Instead, it directly aligns LLMs with human preferences using the Bradley-Terry comparison — a method that involves the probability ranking of n responses generated by the LLM, ensuring they are consistent with human preference rankings.

\paragraph{Data Curation for LLM Alignments.} Many approaches have been devised to curate high-quality, instruction-following datasets to fine-tune LLMs~\citep{wang2022self,wang2023far,alpaca,vicuna2023,peng2023instruction}. For instance, the study by LIMA~\citep{zhou2023lima} underscores that even a limited set of carefully curated and high-quality examples can be utilized to fine-tune a strong pre-trained language model, enabling it to deliver competitive results across a diverse array of prompts. Similarly, \citet{wei2023instructiongpt} introduced a versatile and straightforward data selector designed to autonomously curate a subset from the original fine-tuning dataset, adhering to specific principles for training vision-language models. While these strategies converge on the shared objective of data curation for LLM fine-tuning, our approach is uniquely centered on data curation for PPO training. This strategy diverges fundamentally from others that emphasize the SFT stage, thereby addressing a distinct problem.

% - RLHF
% - PRO - DPO - RAFT - Rejection sampling
% - score normalization

\section{Conclusion}
\label{sec:conclusion}

In this report, we identified and analyzied critical impediments in RLHF training of LLMs, namely reward hacking and catastrophic forgetting. These issues emerge due to the variances in learned reward score distributions and the over-optimization of specific training examples, resulting in instabilities in RLHF training. To alleviate these issues, we introduced the \textit{Advantage Model} and \textit{Selective Rehearsal}—innovative strategies formulated to stabilize the RLHF training process. The Advantage Model aims to maintain balanced reward score distributions across diverse categories and examples, thereby averting complications arising from reward hacking. On the other hand, Selective Rehearsal selectively identifies optimal examples for PPO training, ptimal examples for PPO training, encouraging the retention of crucial knowledge from the SFT stage, and preventing the depreciation of performance over time. Empirical analyses conducted on a range of datasets substantiated the efficacy of our proposed techniques, which not only enhanced stability in RLHF training but also led to improved reward scores and win rates the SFT models.


\bibliography{am}
\bibliographystyle{iclr2021_conference}

% \appendix
% \section{Appendix}

\end{document}